# Mpox Narrative on Instagram: A Labeled Multilingual Dataset of Instagram Posts on Mpox for Sentiment, Hate Speech, and Anxiety Analysis

Nirmalya Thakur
Department of Electrical Engineering and Computer Science
South Dakota School of Mines and Technology
Rapid City, SD 57701, USA
nirmalya.thakur@sdsmt.edu

*Abstract*—The world is currently experiencing an outbreak of mpox, which has been declared a Public Health Emergency of International Concern by WHO. During recent virus outbreaks, social media platforms have played a crucial role in keeping the global population informed and updated regarding various topics. As a result, in the last few years, researchers from different disciplines have focused on the development of social media datasets related to different virus outbreaks, as such datasets serve as a rich data resource for the investigation of a wide range of research questions. No prior work in this field has focused on the development of a dataset of Instagram posts about the mpox outbreak. The work presented in this paper aims to address this research gap and makes two scientific contributions to this field. First, it presents a multilingual dataset of 60,127 Instagram posts about mpox, published between July 23, 2022, and September 5, 2024. The dataset is available at https://dx.doi.org/10.21227/7fvc-y093 and contains Instagram posts about mpox in 52 languages. For each of these posts, the Post ID, Post Description, Date of publication, language, and translated version of the post (translation to English was performed using the Google Translate API) are presented as separate attributes in the dataset. After developing this dataset, sentiment analysis, hate speech detection, and anxiety or stress detection were also performed. This process included classifying each post into (i) one of the fine-grain sentiment classes, i.e., fear, surprise, joy, sadness, anger, disgust, or neutral, (ii) hate or not hate, and (iii) anxiety/stress detected or no anxiety/stress detected. These results are presented as separate attributes in the dataset for the training and testing of machine learning algorithms for sentiment, hate speech, and anxiety or stress detection, as well as for other applications. Second, this paper also presents the results of performing sentiment analysis, hate speech analysis, and anxiety or stress analysis. The variation of the sentiment classes - fear, surprise, joy, sadness, anger, disgust, and neutral were observed to be 27.95%, 2.57%, 8.69%, 5.94%, 2.69%, 1.53%, and 50.64%, respectively. In terms of hate speech detection, 95.75% of the posts did not contain hate, and the remaining 4.25% contained hate. Finally, 72.05% of the posts did not indicate any anxiety/stress, and the remaining 27.95% of the posts represented some form of anxiety/stress.

## I. Introduction

The global resurgence of monkeypox (mpox), caused by the monkeypox virus (MPXV), a zoonotic orthopox virus, remains a pressing public health issue. First identified in 1958 during an outbreak in captive monkeys in Denmark, MPXV was subsequently found to affect humans, with the first case recorded in the Democratic Republic of the Congo (DRC) in 1970 [1, 2]. For decades, mpox was largely confined to Central and West Africa, particularly in the DRC, which has consistently reported the majority of cases [3]. However, recent outbreaks, especially in 2022 and 2024, have elevated mpox to a global concern. Historically, Clade I of the virus, which is more prevalent in Central Africa, has exhibited a higher mortality rate of 10.6%, while Clade IIb, more commonly associated with recent outbreaks, has a lower fatality rate of approximately 3.6% [4].

Beyond Africa, the first significant spread of mpox occurred in 2003, with 47 cases reported in the United States, which were probably linked to the import of infected animals from Ghana [5,6]. Additional outbreaks in Israel and Singapore between 2018 and 2019 were attributed to travelers from Nigeria [7]. The 2022 outbreak, however, was a global turning point, with 99,518 cases reported in 115 regions that did not historically report mpox [8]. Initially, men who have sex with men (MSM) were disproportionately affected [9,10]; subsequent outbreaks have demonstrated that the virus impacts a broader demographic, including children and women. In the DRC, children under 15 account for approximately 66% of cases and more than 82% of deaths due to mpox [11-13].

At the time of writing this paper, there is an ongoing outbreak of mpox. On August 14, 2024, the WHO Director-General declared mpox a Public Health Emergency of International Concern (PHEIC) [14]. The 2024 outbreak has had a more pronounced global impact, particularly in the DRC, where the majority of cases and deaths have occurred. As of August 2024, the African continent has reported over 18,000 cases and 541 deaths, of which children under 15 make up the majority of fatalities [15,16]. This outbreak, driven by Clade IIb, has been exacerbated by inadequate testing and limited vaccine availability. While vaccines such as JYNNEOS, MVA-BN, and LC16 offer some protection, logistical challenges have severely restricted their distribution, particularly in conflict-ridden regions [16].

Infectious disease outbreaks have been a persistent threat to humanity. In the last few years, the usage of social media platforms has skyrocketed, as such platforms serve as virtual communities where people can connect seamlessly with each other [17]. By utilizing concepts of data mining, data analysis, and natural language processing, the patterns of information seeking and sharing on social media platforms during virus outbreaks can be collected [18]. This data is beneficial in understanding multimodal characteristics of content creation and dissemination on social media, which further helps to identify preventive strategies and relevant policies as applicable to public health [19,20]. In view of the various virus outbreaks that have occurred in the last few years, syndromic surveillance via social media, which involves analyzing online content pertaining to public health, is becoming more important than ever [21, 22]. Therefore, the development of datasets of posts from social media platforms such as Twitter, Instagram, Facebook, YouTube, and TikTok, just to name a few, has proven to be highly crucial and valuable for the investigation of a wide range of interdisciplinary research questions related to virus outbreaks and related matters [23].

Of these social media platforms, Instagram stands out as a globally popular social media platform. Instagram has 2.4 billion users on a global scale with India leading as the country with the largest audience of 362 million users. India is followed by the United States and Brazil, whose user counts are 169 million and 134 million, respectively [24]. Brazil is followed by Indonesia, Turkey, Japan, and other countries. Brunei has the highest proportion of users per capita, amounting to 92% of Instagram users in the population. Brunei is followed by Guam and the Cayman Islands, with user penetration rates of 79.2% and 78.8%, respectively [24]. In early 2024, Instagram surpassed the milestone of 2 billion users. This milestone was achieved by Instagram in 11.2 years, which is faster as compared to multiple other social media platforms, for example, Facebook (reached two billion users in 13.3 years) and YouTube (reached 2 billion users in 14 years) [25]. In the United States, Instagram is used by a significant number of social media users and is the third most visited social media site after Facebook and Pinterest [26]. With regards to social media use, 57% of Gen Z users are on Instagram [27], and it has more female users in the United States [28]. In 2023, approximately 80% of marketers worldwide were using Instagram to promote their products and services, which made Instagram rank as the second most used advertising platform after Facebook. In addition to the above, Instagram is the second most accessed social media platform in the United States and accounts for 15.85% of social media visit penetration across desktops, mobiles, and tablets [29]. Despite the global popularity of Instagram, there is still very little research that focuses on the mining and analysis of posts on Instagram related to virus outbreaks. The increasing cases of mpox, along with the measures taken by multiple countries, have led to a tremendous increase in online conversations about MPXV on social media platforms such as Instagram. A recent study highlighted that medical professionals are building their presence as influencers by sharing content related to mpox on social media [30].

No prior work in this field has focused on the development of a dataset of posts on Instagram about the ongoing mpox outbreak. Furthermore, no prior work has presented the analysis of Instagram posts about mpox to detect sentiment, hate, and anxiety. Addressing these major research gaps serves as the main motivation for this work. The rest of this paper is structured as follows. Section II presents a review of recent work in this field. Section III discusses the methodology that was followed to develop this dataset and for performing the data analysis studies. The results are presented in Section IV. Section V concludes the paper and outlines the scope for future work in this field.

## II. LITERATURE REVIEW

This Section presents a review of recent works in this field. In Section II.A a review of recent works related to the development of social media datasets is presented. Section II.B presents a review of recent works related to the analysis of social media posts about mpox.

### A. Review of Recent Works related to the Development of Social Media Datasets

In the last decade and a half, conversations on social media platforms have focused on a wide range of topics, such as virus outbreaks, public health, global concerns, entertainment, politics, sports, fitness, finance, religion, and technology, just to name a few [31,32]. Therefore, the mining of social media posts to develop datasets has attracted the attention of researchers from different disciplines, such as Big Data, Data Mining, and Natural Language Processing.

These datasets have been pivotal for the scientific community in understanding the conversation patterns and the information-seeking behaviors exhibited by the general public related to various topics on different social media platforms. Among these datasets, some recent ones are datasets on hate speech [33], the European migration crisis [34], natural disasters [35], misogynistic language [36], and offensive language [37]. In addition to this, social media datasets have also focused on wide-ranging topics such as civil unrest [38], exoskeletons [39], the effectiveness of hydroxychloroquine (HCQ) for COVID-19 treatment [40], pregnancy [41], measles [42], drug-related knowledge [43], a tornado over Pennsylvania [44], white supremacy [45], Sundanese cultures [46], vaccines [47] and social movements, such as Black Lives Matter [48].

During the COVID-19 pandemic, multiple social media datasets related to the pandemic were developed. These include datasets of social media posts about COVID-19 in Spanish [49], Bengali [50], English [51], Arabic [52], German [53], and French [54]. The development of social media datasets has also included mining of social media posts related to trending topics and hashtags such as #IndonesiaHumanRightsSOS [55], #Blackwomanhood [56], #MarchForBlackWomen [57], #BlackTheory [58], #DuragFest [59], #BringBackOurInternet [60], #WOCAffirmation [61], #AskTimothy [62], #WITBragDay [63], #preuambicio [64], #MiPrimerRecuerdoFeminista [65], #RoeOverturned [66], #SaveKPK [67], #nowplaying [68], #Election2020 [69], and "I Voted For Trump" [70].

These datasets not only provided valuable insights about various topics but were also helpful for the investigation of a wide range of research questions associated with these topics. For instance, the dataset on drug-related knowledge [43] was utilized to track mentions of medications [71], conversations about opioids [72], discussions about birth defects [73], and drug abuse [74] on social media. Researchers have also used this dataset to develop methodologies for detecting breast cancer cohorts from social media data [75], identifying specific drug mentions [76], and online conversations related to adverse drug reactions (ADRs) of marketed drugs [77]. Likewise, the dataset of social media posts about HCQ as a treatment for COVID-19 [40] was used for a wide range of applications such as stance detection [78], misinformation analysis [79], and fake news detection [80], in the context of the public discourse on social media platforms. This dataset was also used for evaluating public perceptions related to using off-label medications for COVID-19 [81] and HCQ as a treatment for COVID-19 [82].

Despite the development of multiple social media datasets, two research gaps still remain. First, none of these datasets focus on the ongoing outbreak of mpox. Second, most of these datasets represent collections of posts from social media platforms other than Instagram, for instance, Twitter, Facebook, YouTube, and TikTok.

*B. Review of Recent Works related to the analysis of Social Media Posts about mpox*

In the last few months, multiple studies have analyzed public sentiment, views, and perspectives toward the mpox outbreak using different social media datasets. These studies highlight different trends, such as the emotions prevalent in discussions, the impact of misinformation, and the stigmatization of certain communities.

Ng et al. [83] examined the public reactions to the mpox outbreak through sentiment analysis of 352,182 tweets that mentioned mpox published between May 6, 2022, and July 23, 2022. Contraire et al. [84] analyzed tweets about mpox published between May 1, 2022, and July 23, 2022. The findings showed that 48,330 of these tweets were posted by individuals from the LGBTQ+ community or their advocates, and the primary sentiment expressed in those tweets was fear or sadness. D'souza et al. [85] analyzed 70,832 tweets with #monkeypox and #LGBTQ+, published between May 1, 2022, and September 7, 2022. Their work showed that mpox-related stigma and misinformation increased online hatred against the LGBTQ+ community on Twitter. The study by Knudsen et al. [86] focused on performing misinformation analysis on social media in the context of mpox. The authors studied tweets published between May 18, 2022, and September 19, 2022. The results showed that 82% of the analyzed tweets contained one or more forms of misinformation about mpox. Zuhanda et al.'s work [87] showed that fear was the dominant sentiment expressed in social media posts about mpox. The authors analyzed 5000 tweets published on August 5, 2022, for their study.

Iparraguirre-Villanueva et al. [88] focused on developing a novel approach for performing sentiment analysis of social media posts about mpox. The model proposed by the authors used a combination of CNN and LSTM and achieved an overall accuracy of 83%. The work of Bengesi et al. [89] also had a similar focus. They used TextBlob, SVM, and concepts of lemmatization and vectorization to develop a sentiment analysis model, which achieved an overall accuracy of 93.48%. Sv et al. [90] studied 556,403 tweets about mpox posted between June 1, 2022, and June 25, 2022, for performing sentiment analysis. Their study reported that 41.6% of the tweets were neutral, 28.82% were positive, and 23.01% were negative. A similar study was performed by Farahat et al. [91]. In this study, the authors analyzed tweets about mpox published between May 22, 2022, and August 5, 2022. Their study reported that 48% of the tweets were neutral, 37% were positive, and 15% were negative

Despite multiple research works in this area, three research gaps exist. First, all these works have focused on the analysis of tweets about mpox, and none of them have focused on the analysis of posts about mpox on Instagram. Second, the majority of these works involve investigating the variation and patterns of sentiment related to mpox, as expressed on social media. However, neither hate speech detection nor anxiety detection was performed in any of these works. Finally, the data used for all these works are social media posts related to the 2022 global outbreak of mpox.

The work presented in this paper aims to address all the research gaps highlighted in II.A and II.B. The methodology that was followed for the development of the dataset, as well as for data analysis, is explained in Section III.

### III. METHODOLOGY

This Section is divided into two parts. In Section III.A, a theoretical overview of the three models that were used for performing sentiment analysis, hate speech detection, and anxiety or stress detection is presented. Section III.B discusses the step-by-step process that was followed for the development of the dataset and for performing data analysis.

*A. Overview of the Models used for Sentiment Analysis, Hate Speech Detection, and Anxiety or Stress Analysis*

The DistilRoBERTa-based model (*j-hartmann/emotion-english-distilroberta-base*) that was used for sentiment analysis [92] is built on a distilled version of RoBERTa, which is an enhancement of the BERT architecture. The distillation process retains approximately 97% of the original RoBERTa model's accuracy while reducing its size and making it 60% faster, which is particularly beneficial for real-time sentiment analysis [93]. The model uses transformer layers to encode the input text, where each word is represented in context using bidirectional self-attention. Mathematically, the model processes the input as a sequence $x=(x_1,x_2,…,x_n)$ and produces hidden states $h_i$ for each word through a series of layers: $h_i=f_\theta(x_i,x_{<i},x_{>i})$, where $f_\theta$ is the function defined by the transformer's parameters. These hidden states are then passed through a classifier to produce the probability distribution over sentiments. The model excels at classifying text into fine-grain sentiments such as fear, surprise, joy, sadness, anger, disgust, and neutral.

For hate speech detection, the *unitary/toxic-bert* model, which is based on BERT [94], was used. BERT operates by tokenizing the input text and then passing the tokens through a bidirectional transformer encoder. BERT's novelty is centered around its bidirectional characteristics as the model learns contextual representation by considering both sides of a token. This can be represented as $p(y|x)=\text{softmax}(W·\text{BERT}(x)+b)$, where W, x, and b represent the learned weight matrix, input token sequence, and bias term, respectively. The *unitary/toxic-*

*bert* model has been trained on large datasets for the detection of toxic content, including hate speech. It uses a softmax layer for the prediction. In this study, the threshold value for this prediction was used as 0.5. The effective pre-training of BERT on masked language modeling enables it to detect contextual meaning for identifying subtle expressions of toxicity that other models may not be able to detect. The same DistilRoBERTa model used for performing sentiment analysis was also used for performing anxiety or stress detection. The DistilRoBERTa classifies sentences by obtaining contextual embeddings for every word and passing the same through a classification layer. The transformer layers map the input tokens to a latent space, $z_i = \text{LayerNorm}(h_i + \text{FFN}(h_i))$, where $h_i$ are the hidden states from the transformer, and the feed-forward network (FFN) refines the representation. This transformation helps capture the intricate patterns in emotional expressions within the text, which can be further analyzed to detect anxiety or stress.

### B. Steps for Dataset Development and Data Analysis

The dataset was developed by mining Instagram posts that comprised #monkeypox or #mpox and were published between July 23, 2022, and September 5, 2024. During the 2022 global outbreak of mpox, on July 23, 2022, the WHO declared mpox a Global Public Health Emergency [95]. So, this date was selected as the start date for the data mining process. September 5, 2024, was the most recent date at the time of writing of this paper. A program was written in Python 3.11.5 for the development of this dataset and the data mining of the relevant Instagram posts, i.e., the posts that contained #monkeypox or #mpox and were published between July 23, 2022, and September 5, 2024, was performed by connecting to the Instagram API [96]. The flowchart shown in Figure 1 outlines the step-by-step process that was followed for the development of this dataset. All the Instagram posts that were collected during this data mining process were publicly available on Instagram and did not require a user to log in to Instagram to view the same (at the time of writing of this paper). After performing data mining, the Google Translate API was used to translate the posts that were published in a language other than English to English. This was an important step as the models used for sentiment analysis, hate speech detection, and anxiety or stress detection are pre-trained on English datasets. To initialize the translation process, the program loaded the credentials for Google Cloud services and set up clients to interact with the Google Translate API. This setup allowed the program to efficiently handle translations within the pipeline, ensuring that Instagram posts from different languages can be processed. However, before passing a post to the translation function called by the Google Translate API, the program detected the language of a post. If the language was English, then the program directly updated the value for the translated post description using the value for the post description to avoid an unnecessary call to the translation function.

However, if the language of a post was not English, then the translation function was called, which used the Google Translate API to translate that post to English. This output was then used to update the value for the translated post description.

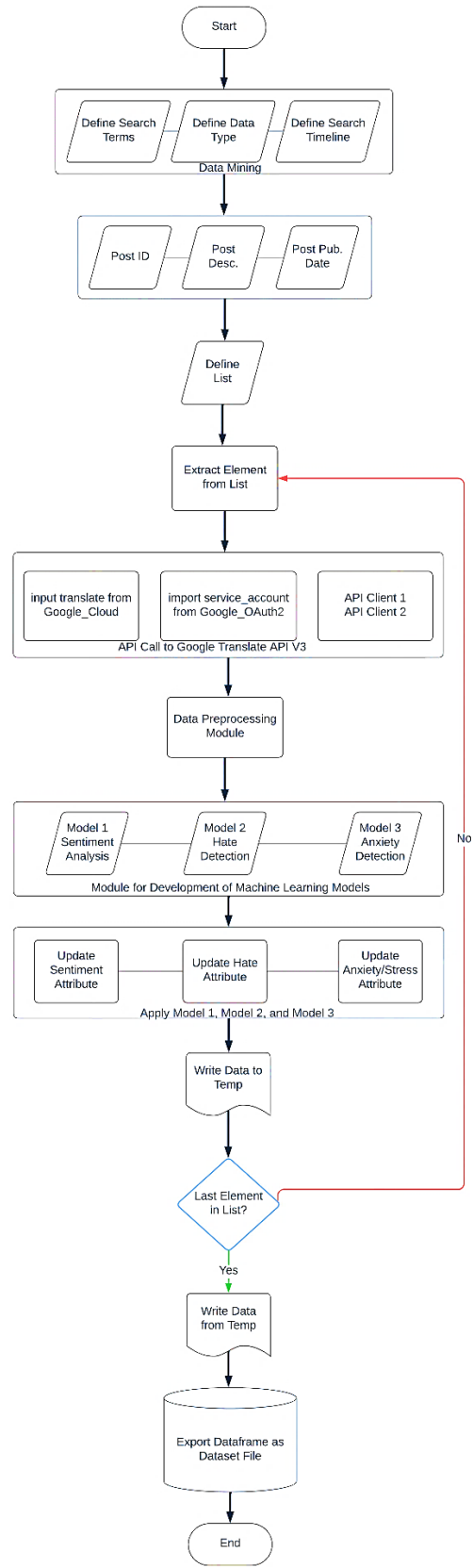

Figure 1. A flowchart that shows the step-by-step process that was followed for the development of this dataset

The program also included multiple forms of error handling to avoid termination of the execution due to any issues associated with the API call. After performing the translation, the next step was data preprocessing. The data preprocessing steps included removing special characters, removing user mentions, removing hashtags, removing punctuation, detecting English words, removing stop words, and removing digits from the posts. During this process, the terms "mpox", "monkeypox", and "monkey pox" were not removed to retain the contextual information. The cleaned and preprocessed text was then analyzed by the models for sentiment analysis, hate speech detection, and stress or anxiety detection, respectively. The model for sentiment analysis classified each Instagram post into one of the fine-grain sentiment classes, i.e., fear, surprise, joy, sadness, anger, disgust, or neutral. The model for hate speech detection worked as a binary classifier and classified each Instagram post as Hate or Not Hate. Finally, the model for anxiety or stress detection also worked as a binary classifier and classified each post as Stress/Anxiety Detected or No Stress/Anxiety Detected. These results were stored as separate attributes in the dataset. The results of sentiment analysis, hate speech detection, and anxiety detection in these attributes were manually verified, and any errors in classification were corrected prior to performing the data analysis.

## IV. RESULTS AND DISCUSSIONS

This Section presents the results. The dataset that was developed is available on IEEE Dataport at https://dx.doi.org/10.21227/7fvc-y093. The dataset contains 60,127 Instagram posts about mpox published between July 23, 2022, and September 5, 2024, in 52 different languages. The distinct languages are English, Portuguese, Indonesian, Spanish, Korean, French, Hindi, Finnish, Turkish, Italian, German, Tamil, Urdu, Thai, Arabic, Persian, Tagalog, Dutch, Catalan, Bengali, Marathi, Malayalam, Swahili, Afrikaans, Panjabi, Gujarati, Somali, Lithuanian, Norwegian, Estonian, Swedish, Telugu, Russian, Danish, Slovak, Japanese, Kannada, Polish, Vietnamese, Hebrew, Romanian, Nepali, Czech, Modern Greek, Albanian, Croatian, Slovenian, Bulgarian, Ukrainian, Welsh, Hungarian, and Latvian. The data description of this dataset is shown in Table 1. As stated in Table 1, this dataset presents the IDs of these posts instead of the URLs of these posts to prevent direct identification of the Instagram users who published these posts. For any post on Instagram, if the Post ID is known, it can be substituted in "PostIDhere" in the generic representation of an Instagram URL: https://www.instagram.com/p/PostIDhere/, to obtain the complete URL of that post. The top 20 languages and the number of times Instagram posts are present in these languages are shown in Table 2.

Table 1: Data Description of the Developed Dataset

| Attribute Name | Attribute Description |
|---|---|
| Post ID | Unique ID of each Instagram post |
| Post Description | Complete description of each post in the language in which it was originally published on Instagram. |
| Date | Date of publication in MM/DD/YYYY format |
| Language | Language of the post as detected using the Google Translate API |
| Translated Post Description | Translated version of the post description. All posts which were not in English were translated into English using the Google Translate API. No language translation was performed for English posts. |
| Sentiment | Results of sentiment analysis where each post was classified into one of the sentiment classes: fear, surprise, joy, sadness, anger, disgust, and neutral |
| Hate | Results of hate speech detection where each post was classified as hate or not hate |
| Anxiety or Stress | Results of anxiety or stress detection where each post was classified as stress/anxiety detected or no stress/anxiety detected. |

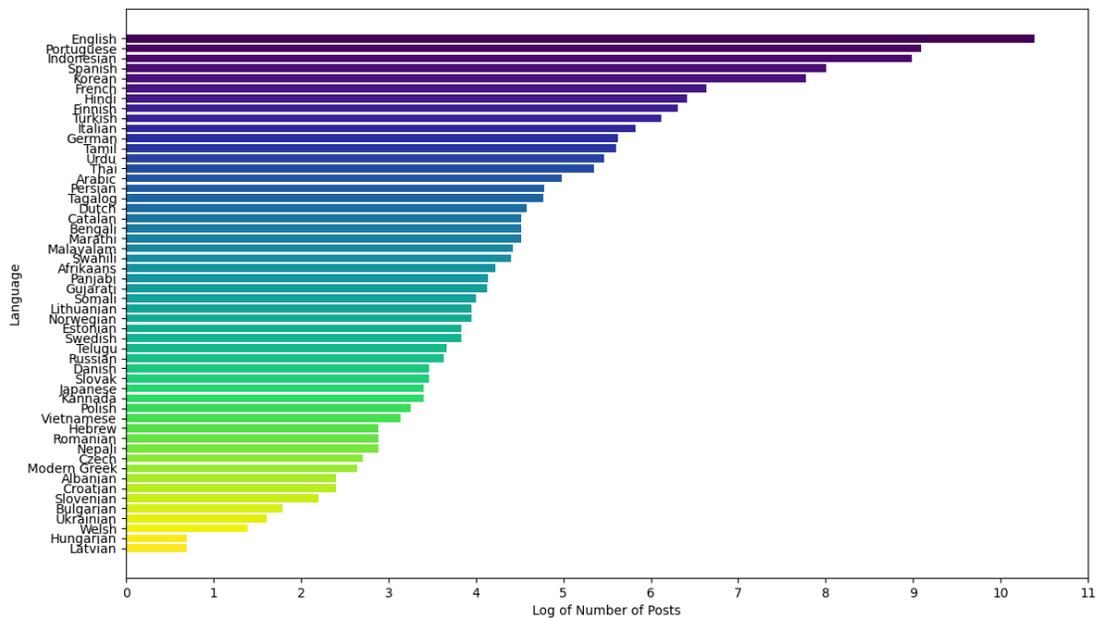

Figure 2: A bar graph that represents the visualization of the log of number of posts per language in this dataset

Table 2: Top 20 languages present in the dataset and their respective frequencies

| Language | Frequency |
|---|---|
| English | 32337 |
| Portuguese | 8926 |
| Indonesian | 7991 |
| Spanish | 3015 |
| Korean | 2390 |
| French | 766 |
| Hindi | 610 |
| Finnish | 553 |
| Turkish | 455 |
| Italian | 339 |
| German | 276 |
| Tamil | 272 |
| Urdu | 236 |
| Thai | 212 |
| Arabic | 146 |
| Persian | 119 |
| Tagalog | 118 |
| Dutch | 98 |
| Catalan | 92 |
| Bengali | 92 |

Figure 2 shows a plot between all the distinct languages and logarithmic values of their frequencies. The results of sentiment analysis, hate speech detection, and anxiety or stress detection are presented in Figures 3, 4, and 5, respectively. From these figures, it can be seen that (i) the variation of the fine-grain sentiment classes: fear, surprise, joy, sadness, anger, disgust, and neutral were 27.95%, 2.57%, 8.69%, 5.94%, 2.69%, 1.53%, and 50.64%, respectively, (ii) 95.75% of these posts did not contain hate and the remaining 4.25% of the posts contained hate, and (iii) in 72.05% of the posts no anxiety or stress was detected and the remaining 27.95% of the posts represented some form of anxiety or stress.

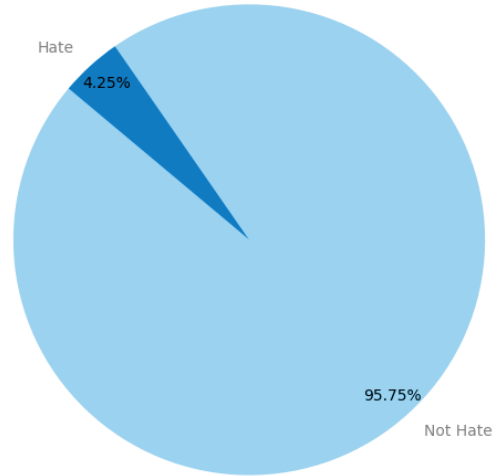

Figure 4 A pie chart that represents the results of hate speech detection

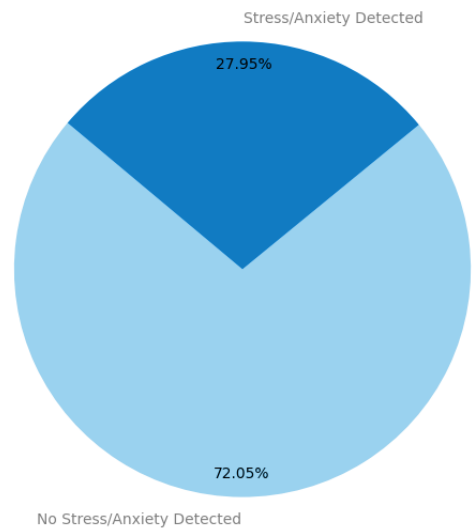

Figure 5. A pie chart that represents the results of anxiety or stress detection

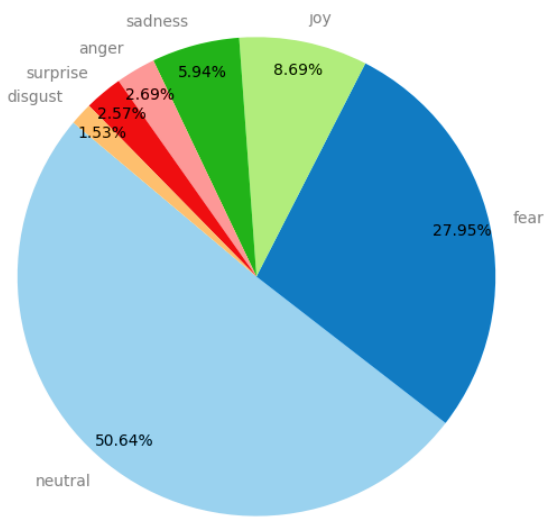

Figure 3: A pie chart that represents the results of sentiment analysis

A discussion to outline the compliance of this dataset with the FAIR principles of scientific data management [97] is presented next. FAIR stands for Findability, Accessibility, Interoperability, and Reusability. The FAIR principles outline essential considerations in data publishing, which are designed to support both manual and automated processes for data submission, discovery, access, collaboration, and reuse. Adherence to these principles can vary and evolve as data publishers enhance their practices towards greater compliance with FAIR. It is important to note that the FAIR principles are not a standard. Instead, they offer guidelines that allow data publishers and managers to reflect on the extent to which their decisions respect the principles of Findability, Accessibility, Interoperability, and Reusability [97]. Multiple prior works related to dataset development have discussed how the developed datasets adhere to the FAIR principles. Examples of datasets that comply with the FAIR principles include - the human metabolome database [98], the WikiPathways dataset [99], a dataset of tweets about COVID-19 [100], the computational 2D materials database (C2DB) [101], the open reaction database [102], RCSB Protein Data Bank [103], and PHI-base [104]. This dataset meets the FAIR principles effectively. It is findable by a unique and permanent DOI provided by IEEE Dataport, ensuring it can be located by researchers across disciplines. It is accessible globally via this DOI, provided there is internet connectivity, and the device used to access the internet is functional. The dataset is interoperable, as the data in this dataset is available in a standard format (.xlsx file) that can be downloaded, read, and analyzed across different computer systems, frameworks, and applications. Lastly, this dataset satisfies the reusability property as the data can be re-used any number of times for the study and investigation of different research questions that focus on the analysis of Instagram posts related to mpox.

This paper has a few limitations. First, even though it is not stated in the description of *j-hartmann/emotion-english-distilroberta-base* [92], it was observed that this model could process up to 512 characters. So, to address this issue, the first 512 characters from the preprocessed version of each Instagram post were passed to this model for sentiment analysis. For consistency, the same data was passed to the models for hate speech detection and anxiety or stress detection. Second, the Google Translate API was used to translate all the Instagram posts which were not in English. However, these translations were not verified by native speakers of those languages for correctness. Third, as stated in Section III, the results of sentiment analysis, hate speech detection, and anxiety detection presented in the attributes of the developed dataset were manually verified, and any errors in classification were corrected. However, there may be human errors associated with the manual verification process [105]. Finally, the results of sentiment analysis, hate speech analysis, and anxiety or stress analysis, as discussed in this paper, are based on the data present in this dataset. As conversations on Instagram keep evolving on a frequent basis, it is possible that if new data related to mpox posts on Instagram is collected in the future and sentiment analysis, hate speech analysis, and anxiety or stress analysis is performed on the same, the results obtained from such a study may vary from the results presented in this paper.

## V. Conclusion

In the modern-day Internet of Everything lifestyle, people use social media more than ever. By utilizing concepts of data mining, data analysis, and natural language processing, significant health information can be retrieved from social media platforms. During virus outbreaks of the recent past, social media platforms have been invaluable in uncovering insights related to the patterns of public views, perspectives, and reactions related to the outbreaks. Therefore, the development of social media datasets has attracted the attention of researchers from different disciplines, and multiple social media datasets related to COVID-19 have been developed in the last couple of years or so. The world is currently experiencing an ongoing outbreak of mpox, which has been declared a Public Health Emergency of International Concern by WHO. No prior work in this field has focused on the development of a dataset of Instagram posts about mpox or performing analysis of such posts on Instagram to detect sentiment, hate, and anxiety.

The work presented in this paper addresses these research gaps. It presents a dataset of 60,127 Instagram posts about mpox, published between July 23, 2022, and September 5, 2024. This is a multilingual dataset that contains posts in 52 different languages. There are different attributes in this dataset that present specific information about the posts. These attributes are Post ID, Post description, Date, Language, Translated Post Description, Sentiment, Hate, and Anxiety or Stress. The model for sentiment analysis classified each Instagram post into one of the fine-grain sentiment classes, i.e., fear, surprise, joy, sadness, anger, disgust, or neutral. The model for hate speech detection worked as a binary classifier and classified each Instagram post as Hate or Not Hate. Finally, the model for stress or anxiety detection also worked as a binary classifier and classified each post as Stress/Anxiety Detected or No Stress/Anxiety Detected. These results per post were stored in the last three attributes of the dataset. The dataset complies with the FAIR principles of scientific data management. The paper also presents the findings of performing sentiment analysis, hate speech detection, and anxiety or stress detection of these Instagram posts. The variation of the fine-grain sentiment classes: fear, surprise, joy, sadness, anger, disgust, and neutral were 27.95%, 2.57%, 8.69%, 5.94%, 2.69%, 1.53%, and 50.64%, respectively. In terms of hate detection, 95.75% of these posts did not contain hate, and the remaining 4.25% contained hate. Finally, 72.05% of the posts did not indicate any anxiety or stress, and the remaining 27.95% of the posts represented some form of anxiety or stress. Future work would involve performing topic modeling using this dataset to identify the specific topics and trends in the context of Instagram posts about mpox.

## Conflicts of Interest

The author declares no conflicts of interest.